\documentclass[letterpaper]{article} % DO NOT CHANGE THIS
\usepackage{aaai2026}  % Changed from [submission] for arXiv compatibility
\usepackage{times}  % DO NOT CHANGE THIS
\usepackage{helvet}  % DO NOT CHANGE THIS
\usepackage{courier}  % DO NOT CHANGE THIS
\usepackage[hyphens]{url}  % DO NOT CHANGE THIS
\usepackage{graphicx} % DO NOT CHANGE THIS
\usepackage{natbib}  % DO NOT CHANGE THIS AND DO NOT ADD ANY OPTIONS TO IT
\usepackage{caption} % DO NOT CHANGE THIS AND DO NOT ADD ANY OPTIONS TO IT
\usepackage{amsmath}
\usepackage{amssymb}
\usepackage{amsthm}
\usepackage{subcaption}
\usepackage{booktabs}
\usepackage{xcolor}
\usepackage{colortbl}
\usepackage{pifont}

\usepackage{pgfplots}
\usepackage{graphicx}
\usepackage{caption}
\usepackage{subcaption}
\usetikzlibrary{3d, arrows.meta, positioning}
\pgfplotsset{compat=newest}

\newcommand{\xmark}{\ding{55}}  % ✗
\makeatletter
\@ifundefined{isChecklistMainFile}{
  \newif\ifreproStandalone
  \reproStandalonetrue
}{
  \newif\ifreproStandalone
  \reproStandalonefalse
}
\makeatother

\ifreproStandalone
\usepackage{aaai2026}
\usepackage{times}
\usepackage{helvet}
\usepackage{courier}
\usepackage{xcolor}
\nocopyright

\title{Unconsciously Forget: Mitigating Memorization \\Without Knowing What is being Memorized}
\author{
Er Jin$^{1}$\thanks{Equal contribution},
Yang Zhang$^{2}$\footnotemark[1],
Yongli Mou$^{3}$,
Yanfei Dong$^{4}$,\\
\textbf{Stefan Decker$^{3,5}$,
Kenji Kawaguchi$^{2}$,
Johannes Stegmaier$^{1}$}
}
\affiliations{
$^{1}$Institute of Imaging and Computer Vision, RWTH Aachen University, Aachen, Germany\\
$^{2}$National University of Singapore\\
$^{3}$Department of Computer Science, RWTH Aachen University, Aachen, Germany\\
$^{4}$Independent Researcher\\
$^{5}$Fraunhofer Institute for Applied Information Technology FIT, Sankt Augustin, Germany
}

\begin{document}

\maketitle

\begin{abstract}

Recent advances in generative models have demonstrated an exceptional ability to produce highly realistic images. However, previous studies show that generated images often resemble the training data, and this problem becomes more severe as the model size increases. Memorizing training data can lead to legal challenges, including copyright infringement, violations of portrait rights, and trademark violations. Existing approaches to mitigating memorization mainly focus on manipulating the denoising sampling process to steer image embeddings away from the memorized embedding space or employ unlearning methods that require training on datasets containing specific sets of memorized concepts. However, existing methods often incur substantial computational overhead during sampling, or focus narrowly on removing one or more groups of target concepts, imposing a significant limitation on their scalability. To understand and mitigate these problems, our work, \emph{UniForget}, offers a new perspective on understanding the root cause of memorization. Our work demonstrates that specific parts of the model are responsible for copyrighted content generation. By applying model pruning, we can effectively suppress the probability of generating copyrighted content without targeting specific concepts while preserving the general generative capabilities of the model. Additionally, we show that our approach is both orthogonal and complementary to existing unlearning methods, thereby highlighting its potential to improve current unlearning and de-memorization techniques.

\end{abstract}


\section{Introduction}
\label{sec:intro}

\begin{figure}[t]
    \centering
    \includegraphics[width=\linewidth]{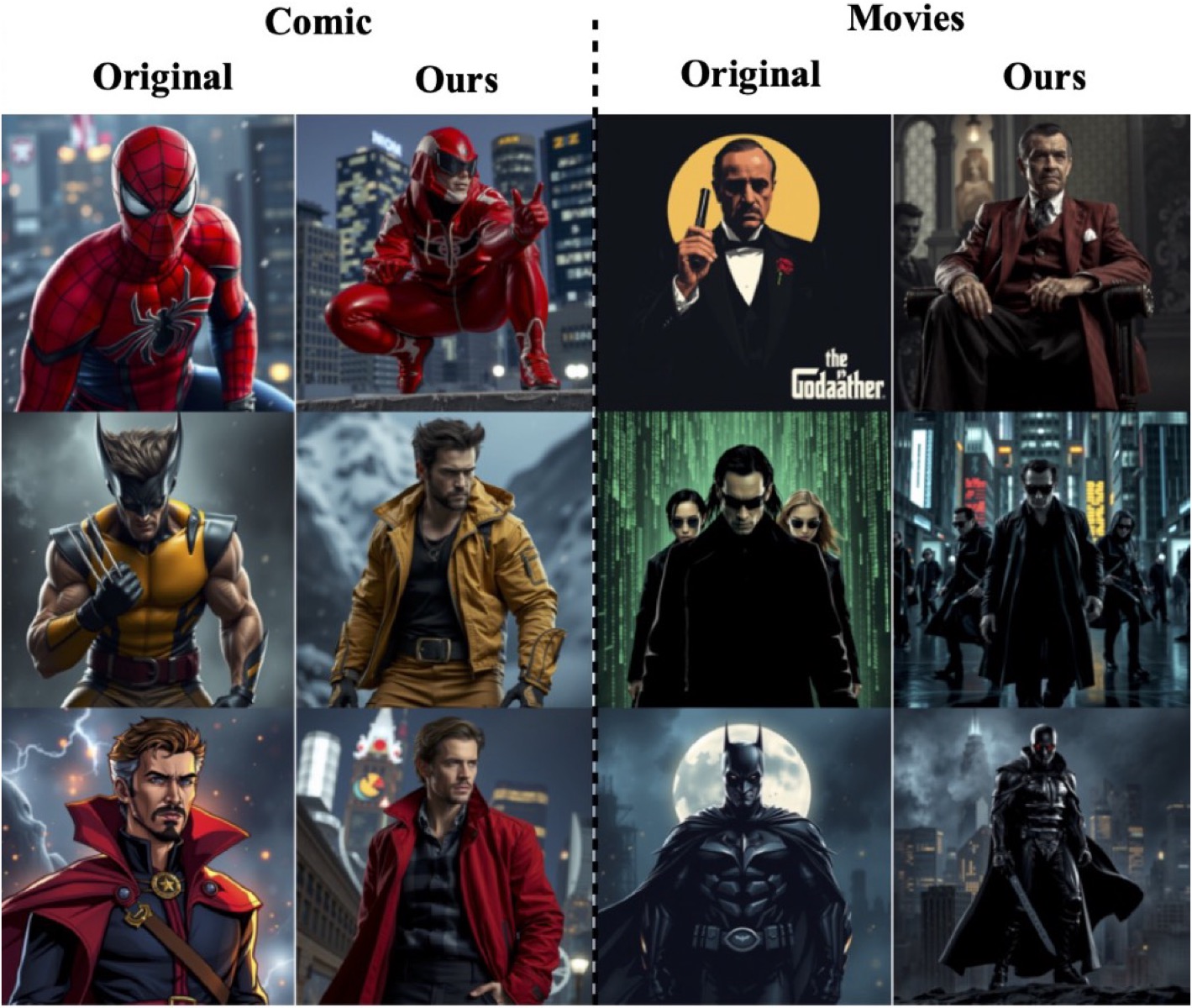}
    \caption{Example images of our proposed method UniForget, which can greatly mitigate copyright infringement problems from FLUX without impairing the generation quality. \textbf{Our method can effectively mitigate the risk of generating over 2000 copyright-sensitive concepts}. Each pair of images is generated with an identical prompt, comparing the original FLUX model with our de-memorized model.}
    \label{fig:teaser}
\end{figure}

\pgfplotsset{
  colormap={softdensity}{
    rgb255(0cm)=(210,250,210)   % pale green (very low density)
    rgb255(1cm)=(255,230,100)   % warm gold
    rgb255(2cm)=(200,60,60)     % soft red (high density)
  }
}

\pgfplotsset{compat=1.18}  % Make sure your PGFPlots version is compatible

\begin{figure*}[t]
\centering

% ---------- Subfigure (a): Before De-memorization ----------
\begin{subfigure}[t]{0.48\textwidth}
\centering
\begin{tikzpicture}[font=\small]
\begin{axis}[
    view={45}{40},
    hide axis,
    domain=-4:4,
    y domain=-4:4,
    colormap name=softdensity,
    samples=40,
    z buffer=sort,
    width=12cm,
    height=8cm,
    axis lines=none,
    enlargelimits=true,
    point meta min=0,
    point meta max=1,
]
    \addplot3[surf, opacity=0.6]{exp(-0.4*(x^2 + y^2))};
    \addplot3[surf, domain=-2.5:-1.5, y domain=-0.5:0.5, samples=20, opacity=0.9]
        {0.9*exp(-25*((x+2)^2 + y^2))};
    \addplot3[surf, domain=1.5:2.5, y domain=-0.5:0.5, samples=20, opacity=0.9]
        {1.3*exp(-25*((x-2)^2 + y^2))};

    \node at (axis cs:0, 0, 0) {\scriptsize \textbf{Neutral Subspace}};
    \node[align=center, text width=3.5cm] at (axis cs:2.8,-0.6, -0.1)
        {\scriptsize \textbf{Subspace of\\Memorized Content}};

    \draw[->, thick, dashed] (axis cs:-3.5,-1.2,0.3) -- (axis cs:-2, -0.1, 0.3);
    \node at (axis cs:-3, -3, 0) {\textbf{$\mathcal{Z}_{N}$}};
    \draw[->, thick, dashed] (axis cs:1, 3.5, 0) -- (axis cs:0.8,1.4,0);

    % ---------- suppression arrows (professional style) ----------
    \draw[->, >=Stealth, line width=1.5pt, color=blue!70]
          (axis cs:-2,0,1.25) -- (axis cs:-2,0,0.45);   % left spike
    \draw[->, >=Stealth, line width=1.5pt, color=blue!70]
          (axis cs: 2,0,1.65) -- (axis cs: 2,0,0.55);   % right spike

    % optional tiny labels
    \node[font=\scriptsize\sffamily, color=red!75!black]
         at (axis cs:-2.15,0,1.3) {suppressing};
    \node[font=\scriptsize\sffamily, color=red!75!black]
          at (axis cs: 1.85,0,1.7) {suppressing};

    \node at (axis cs:-3.5,-1.2,0.3)
      {\includegraphics[width=1cm]{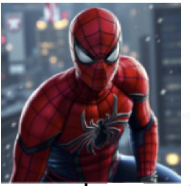}};
    \node at (axis cs:1, 3.5, 0)
      {\includegraphics[width=1cm]{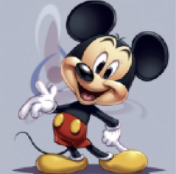}};
\end{axis}
\end{tikzpicture}
\caption*{(a) Before De-memorization}
\end{subfigure}
\hfill
% ---------- Subfigure (b): After De-memorization ----------
\begin{subfigure}[t]{0.48\textwidth}
\centering
\begin{tikzpicture}[font=\small]
\begin{axis}[
    view={45}{40},
    hide axis,
    domain=-4:4,
    y domain=-4:4,
    colormap name=softdensity,
    samples=40,
    z buffer=sort,
    width=12cm,
    height=8cm,
    axis lines=none,
    enlargelimits=true,
    point meta min=0,
    point meta max=1,
]
    \addplot3[surf, opacity=0.6]{exp(-0.4*(x^2 + y^2))};
    \addplot3[surf, domain=-2.5:-1.5, y domain=-0.5:0.5, samples=20, opacity=0.5]
        {0.2*exp(-25*((x+2)^2 + y^2))};
    \addplot3[surf, domain=1.5:2.5, y domain=-0.5:0.5, samples=20, opacity=0.5]
        {0.3*exp(-25*((x-2)^2 + y^2))};

    \node at (axis cs:0, 0, 0) {\scriptsize \textbf{Neutral Subspace}};
    \node[align=center, text width=3.5cm] at (axis cs:2.8,-0.6, -0.1)
        {\scriptsize \textbf{Subspace of\\Memorized Content}};

    \draw[->, thick, dashed] (axis cs:-3.5,-1.2,0.3) -- (axis cs:-2, -0.1, 0.1);
    \node at (axis cs:-3, -3, 0) {\textbf{$\mathcal{Z}_{N}$}};
    \draw[->, thick, dashed] (axis cs:1, 3.5, 0) -- (axis cs:0.8,1.4,-0.1);

    \node at (axis cs:-3.5,-1.2,0.3)
      {\includegraphics[width=1cm]{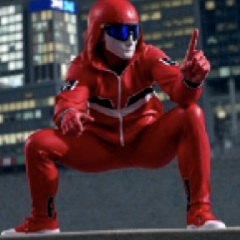}};
    \node at (axis cs:1, 3.5, 0)
      {\includegraphics[width=1cm]{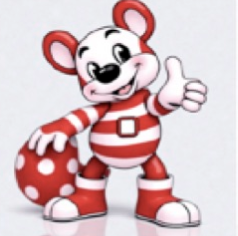}};
\end{axis}
\end{tikzpicture}
\caption*{(b) After De-memorization}
\end{subfigure}

\caption{
    (a) demonstrates the sampling process of an image generative model, which projects from a random noise distribution to a target distribution. Hypothetically, this distribution contains spikes corresponding to the \textbf{subspace of memorized content}. These spikes with exceptionally high probability density indicate that the model tends to remember the projection path to those target spikes when using copyright-sensitive prompts. 
    (b) We aim to largely suppress the probability of these spikes and force the model to project to the neutral embedding space, thereby mitigating the risk of model memorization.}
    \label{fig:hypothesis}
\end{figure*}
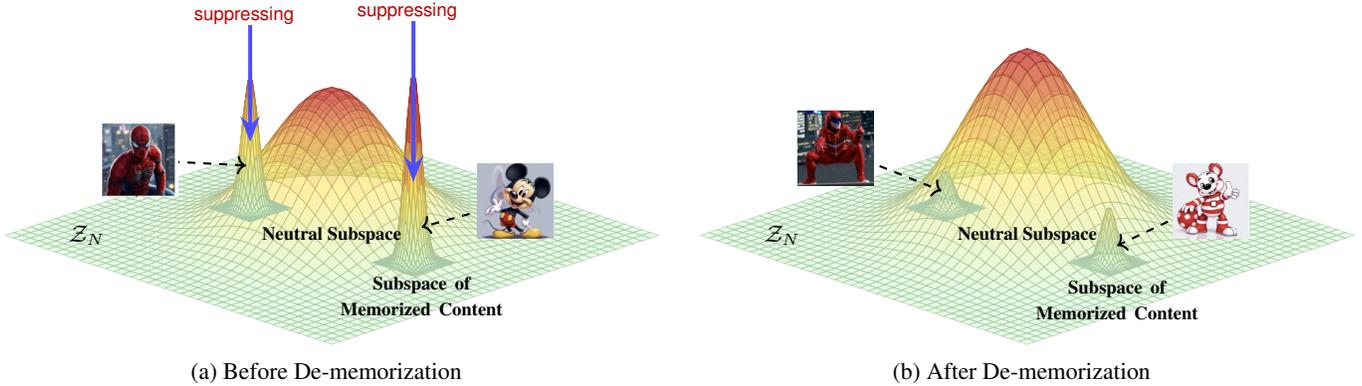

Generative models~\cite{ho2020denoising, nichol2021improved, rombach2022high, podell2023sdxl, peebles2023scalable, blackforestlabs2024} excel in generating high-fidelity images, in part due to large-scale web-crawled image-caption datasets such as LAION and GCC3M~\cite{schuhmann2021laion, schuhmann2022laion, wang2019learning}. Training on these datasets and increasing model size can significantly improve generative performance~\cite{liang2024scaling, tian2024visual, podell2023sdxl}. For example, recent models, such as Stable Diffusion 3 and FLUX, have benefited from these advancements~\cite{blackforestlabs2024, rombach2022high}. However, many studies show that these datasets often contain copyrighted materials, unauthorized portraits, and celebrity likenesses, which can lead to undesirable behavior in the trained model~\cite{sag2023copyright, oratz2024initial}. One of the most critical problems is model memorization, especially the memorization of copyrighted or sensitive content from the training dataset~\cite{somepalli2023understanding}. The widespread adoption of generative models will be significantly hindered if model memorization remains unresolved due to the risk of copyright violations and other potentially irresponsible behavior.

Addressing model memorization remains a critical challenge in generative models. \citet{wen2024detecting} shows the large magnitude of text-conditional noise in diffusion models is indicative of memorization and that perturbing the input embedding can effectively reduce this magnitude, thereby mitigating memorization. Determining the required level of perturbation relies on manually set thresholds, making the approach impractical and leading to inconsistent results. An alternative method employs multiple guidance strategies to steer the denoised latent toward an embedding space that maintains a sufficient distance from the training data while preserving generative quality~\cite{chen2024towards}. However, this method still relies on fixed detection thresholds, which can cause misalignment between text and image generation and introduce unwanted artifacts, ultimately limiting its effectiveness. We observe that most existing methods focus on modifying either prompt tokens or denoised latents to push outputs away from memorized embeddings, increasing complexity and causing noticeable sampling overhead. While these approaches help mitigate memorization, they do not address the root cause: the very existence of a memorized embedding space within the model. 

Other than these de-memorization methods, many unlearning methods~\cite{kumari2023ablating, gandikota2023erasing, buierasing, lu2024mace, gandikota2024unified} attempt to reduce the memorized embedding space by restricting the model's ability to generate images of specific target concepts. However, these methods have significant limitations. As the number of targeted concepts increases, the generative performance of the model tends to deteriorate considerably~\cite{gandikota2024unified}. Consequently, it is often necessary to train separate models for different concepts across various domains in real-world scenarios, which is impractical. In this work, we present \emph{UniForget}, a method that addresses memorization by identifying and removing the specific parts of the model responsible for copyrighted content generation. Our approach is based on two key hypotheses: (1) the embedding space of memorized copyrighted content can be decoupled from neutral features, and (2) a small subset of model parameters is responsible for generating copyrighted content. Our main contributions are as follows:
\begin{itemize}
\item We propose UniForget, a novel approach that mitigates copyrighted memorization without requiring knowledge of specific copyrighted concepts.
\item We validate our hypotheses through comprehensive experiments, demonstrating that our method effectively decouples copyrighted embeddings from neutral content while preserving general generative capabilities.
\item We show that our approach is orthogonal and complementary to existing unlearning methods, opening new avenues for responsible AI development.
\item We conduct extensive evaluations across multiple copyright-sensitive domains, showing significant improvements in reducing memorization while maintaining competitive performance on neutral prompts.
\end{itemize}

\section{Preliminaries and Related Work}

We provide an overview of two foundational preliminaries: flow matching models (Appendix A) and their application to image generation (Appendix B).

\label{sec:related works}

\subsection{Training Data Memorization.} Many recent works~\cite{somepalli2023understanding, carlini2021extracting} have shown that model memorization issues are ubiquitous in modern generative models, including Stable Diffusion~\cite{rombach2022high}, Imagen~\cite{saharia2022photorealistic}, and DALL·E~\cite{ramesh2021zero}. Using the data extraction attack method proposed by \citet{carlini2023extracting} memorized content can be easily regenerated, increasing the risk of generating copyrighted content. This poses a significant limitation on the practical use of these generative models.

\subsection{Sampling Processing Manipulation.} Recent approaches such as Anti-Memorization Guidance (AMG)~\cite{chen2024towards} and the method proposed by \citet{wei2024memorization} demonstrate that current methods can largely mitigate memorization issues by manipulating the guidance or adjusting the input text prompts embeddings. Other manipulation methods, such as FlowEdit~\cite{kulikov2024flowedit} and RF-inversion~\cite{rout2025semantic}, have also demonstrated the potential of utilizing flow-matching editing techniques to modify image embeddings, thereby mitigating memorization issues. All these methods could potentially be applied to remove copyrighted \mbox{concepts}.

\subsection{Concept Removal.} For removing specific concepts, various methods in the unlearning domain, such as Concept Ablation (CA)~\cite{kumari2023ablating}, Erasing Concepts from Diffusion Models (ESD)~\cite{gandikota2023erasing}, and Erasing Adversarial Preservation (EAP)~\cite{buierasing}, effectively eliminate specific target concepts by manipulating cross-attention modules or, they erase the target concepts and replace them with anchor concepts via fine-tuning the attention layers. Other approaches, like MAss Concept Erase (MACE)~\cite{lu2024mace} and Unified Concept Editing (UCE)~\cite{gandikota2024unified}, demonstrate that models can forget up to 100 target concepts without sacrificing generative quality.

In this paper, our work aims to mitigate and understand the problems in an \textbf{orthogonal} manner to all current approaches. By integrating our method with existing techniques, we firmly believe that the problem of memorization of copyrighted content in generative models can be largely resolved.

\begin{figure*}[t]
    \centering
    \includegraphics[width=\linewidth]{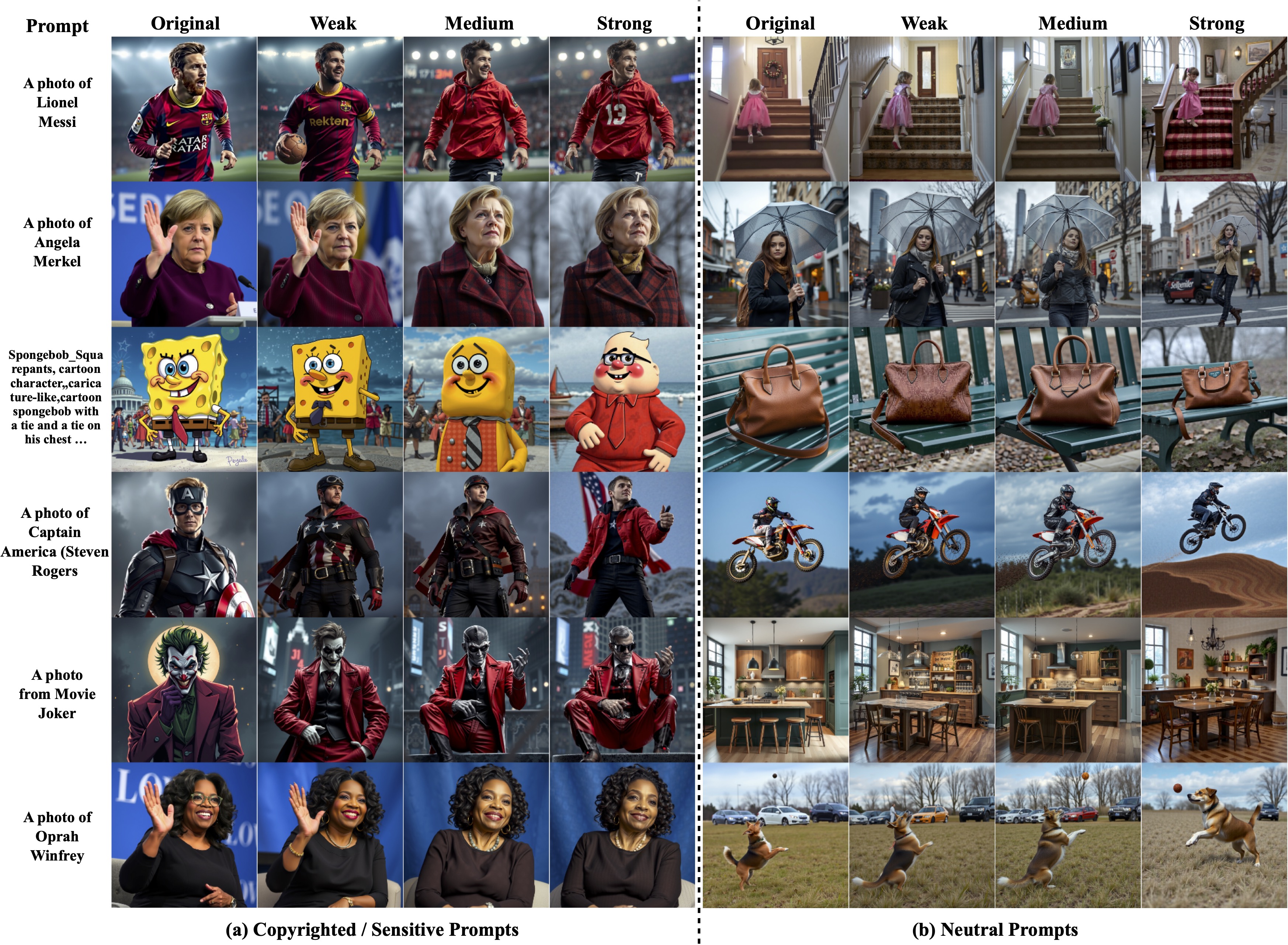}
    \caption{De-memorization results across different categories and levels (weak, medium, strong) using copyright-sensitive prompts and neutral prompts from COCO. We compare our results with the original images generated by FLUX schnell.  (a) Our method effectively de-memorizes copyrighted prompts, ensuring that the generated images preserve the overall structure of the originals while excluding copyrighted elements. (b) shows that for neutral prompts, our method can still largely preserve high fidelity and semantic meaning.}
    \label{fig:demo_results}
\end{figure*}

\section{Methods}\label{sec:methods}

\subsection{Hypotheses on Memorization.} As shown in Figure~\ref{fig:hypothesis}, we assume that the copyrighted embedding is located within a spike in the target distribution. We propose suppressing the spike, thus reducing the likelihood of generating copyrighted content, as depicted in Figure~\ref{fig:hypothesis}. To achieve this, we propose our method based on the following two essential \textbf{hypotheses}: 
(1) \textbf{Decoupled embedding space between neutral and copyrighted features}: The latent embeddings of memorized content can be separated from the embedding space of de-memorized features. Modifying or removing the memorized embeddings does not significantly affect the generation of de-memorized features in the embedding space. (2) \textbf{Selective parameters responsible for memorization}: A small subset of model parameters influences the generation or reproduction of copyrighted content. These parameters cause the generated features to deviate from the intended neutral embedding space.

\subsection{Mitigating Copyright-Sensitive Content} \label{sec:training} \citet{chavhan2024memorized}'s work also demonstrates that pruning can be used to identify the embedding subspace. However, their approach is trained on datasets containing prompts that cause memorization issues. They have not yet explored the subspace of memorized copyrighted content.  Nonetheless, we believe their work is highly valuable and strongly supports our hypothesis as well. Building on \citet{chavhan2024memorized}'s findings and our hypotheses, we propose leveraging pruning methods to de-memorize copyrighted content. Our method \textbf{uses only neutral prompts $\mathcal{C}_n$} for training while deactivating certain parts of the model. We adapt the pruning method from EcoDiff~\cite{zhang2024effortless} with a modified objective function defined as follows:
\begin{equation}\label{eq:pruning_loss}
\resizebox{0.9\columnwidth}{!}{$
     \min_{\theta} \mathbb{E}_{c \sim \mathcal{C}_{n}} \bigg[ 
    \frac{1}{N_l}\left\| \mathbb{F}_u(\mathbf{z}_0, N, c) - \mathbb{F}_{u_{\theta}}(\mathbf{z}_0, N, c)  \right\|_2^2 + \beta \frac{\left\| \theta \right\|_1}{|\theta|} 
    \bigg],$}
\end{equation}

where $N_l$ is the number of elements in the denoised latent, $\mathbb{F}$ denotes the denoising sampling process from Equation Appendix A, $\beta$ is the regularization coefficient and $\left\| \theta \right\|_1$ measures the size of the activated parts of the model and $|\theta|$ is the cardinality, representing the total number of activated parts in the model. The loss function in Equation~\ref{eq:pruning_loss} consists of two main components. The term,
\begin{equation}
    \frac{1}{N_l}\left\| \mathbb{F}_u(\mathbf{z}_0, N, c) - \mathbb{F}_{u_{\theta}}(\mathbf{z}_0, N, c)  \right\|_2^2,
\end{equation}
works as the reconstruction loss that measures the mean square error (MSE) between the denoised latent representation $\mathbf{z}_N$ produced by the original model $u$ and that generated by the pruned model $u_\theta$. This term ensures that the masked model retains the ability to synthesize images conditioned on the neutral prompt $c \in \mathcal{C}_n$, thereby preserving its generative performance. The second term $\beta \frac{\left\| \theta \right\|_1}{|\theta|}$ measures the number of activated parts of model modules. Based on the hypothesis, this term is designed to train the model to remove the subset of the model responsible for generating copyrighted content. To efficiently train the model with limited computational resources, we utilize pruning methods from previous work~\cite{fang2024maskllm,zhang2024effortless} in combination with gradient checkpointing for memory-efficient training~\cite{chen2016training, zhang2024effortless}. In this approach, we apply a learnable mask $\mathcal{M}$ to various neural layers, including feedforward $\mathcal{M}_{\mathrm{ffn}}$, attention $\mathcal{M}_{\mathrm{attn}}$, and normalization layers $\mathcal{M}_{\mathrm{norm}}$. Additionally, we employ the hard discrete relaxation function~\cite{louizos2017learning, zhang2024effortless} to the mask $\mathcal{M}$ during training. The implementation details of the relaxation function are provided in Appendix C. The objective function is modified as follows:
\begin{equation}\label{eq:modified_objective}
\resizebox{0.9\columnwidth}{!}{$
    \min_{\theta} \mathbb{E}_{c \sim \mathcal{C}_{n}} \bigg[
    \frac{1}{N_l}\left\| \mathbb{F}_{u_{\theta}}(\mathbf{z}_0, N, c) - \mathbb{F}_u(\mathbf{z}_0, N, c) \right\|_2^2 
    + \beta \frac{\left\| \hat{\sigma}(\mathcal{M}) \right\|_1 }{|\mathcal{M}|}
    \bigg],$}
\end{equation}

where $\hat{\sigma}$ denotes the relaxed sigmoid function applied to the mask $\mathcal{M}$ as follows:
\begin{equation} \label{eq:relax_sigmoid}
    \mathcal{\hat{M}} = \hat{\sigma}(\mathcal{M})  = \sigma(\mathcal{M} \cdot \gamma + \delta) ,
\end{equation}
where $\gamma$ and $\delta$ denote the regularization terms that ensure smoother training. The default values are $\gamma = 0.4$ and $\delta = 1$.  Additionally, $|\mathcal{M}|$ represents the number of masked elements. The parameter $\beta$ regularizes the number of activated elements in $\mathcal{M}$. If our hypothesis in Methods Section holds, $\beta$ also controls the degree of de-memorization: lower values of $\beta$ result in weaker de-memorization effects. In comparison, higher values lead to stronger suppression of memorization. Therefore, we conduct experiments across three distinct de-memorization settings, weak, medium, and strong, defined by the following hyperparameters: $\beta_{\text{weak}} = 1, \beta_{\text{medium}} = 2, \beta_{\text{strong}} = 5.$ All the parameters $\beta$ are determined through ablation studies, and detailed information is provided in Appendix G.

\begin{figure*}[t]
    \centering
    \hfill
    \begin{subfigure}[b]{0.32\textwidth}
        \centering
        \includegraphics[width=\textwidth]{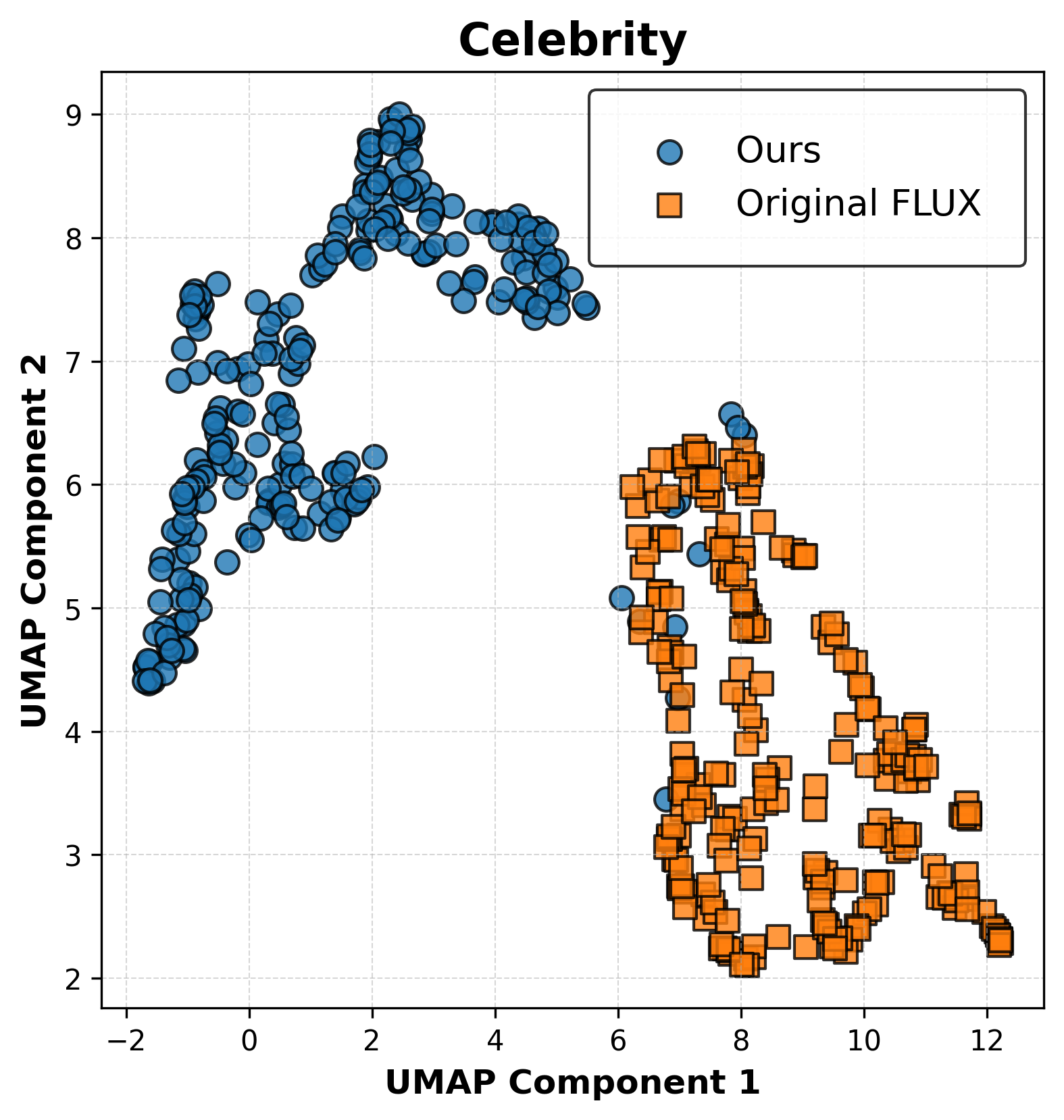}
        \caption{Memorized Concept: Celebrity}
        \label{fig:cele}
    \end{subfigure}
    \begin{subfigure}[b]{0.32\textwidth}
        \centering
        \includegraphics[width=\textwidth]{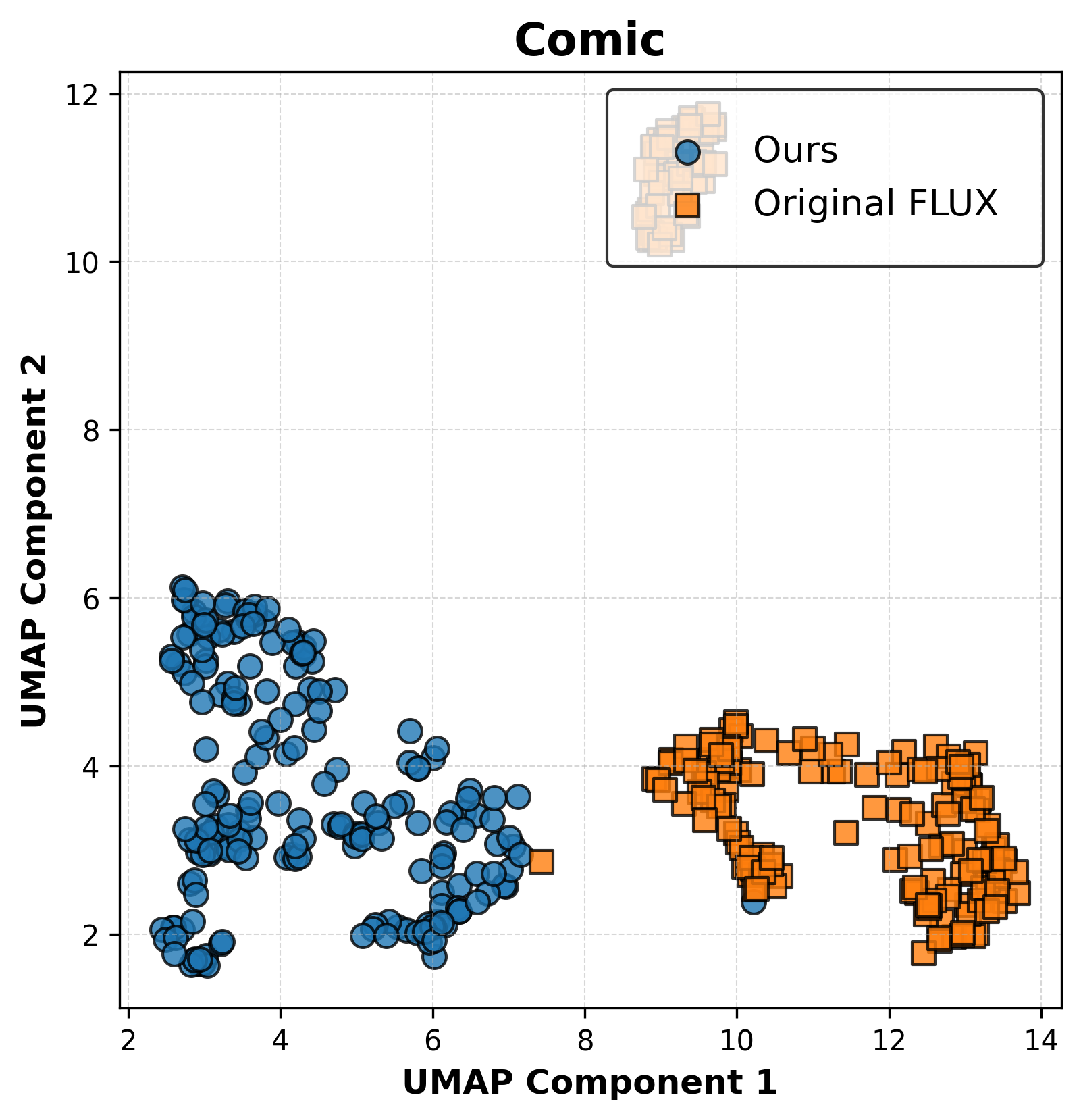}
        \caption{Memorized Concept: Comic}
        \label{fig:comic}
    \end{subfigure}
    \hfill
    \begin{subfigure}[b]{0.32\textwidth}
        \centering
        \includegraphics[width=\textwidth]{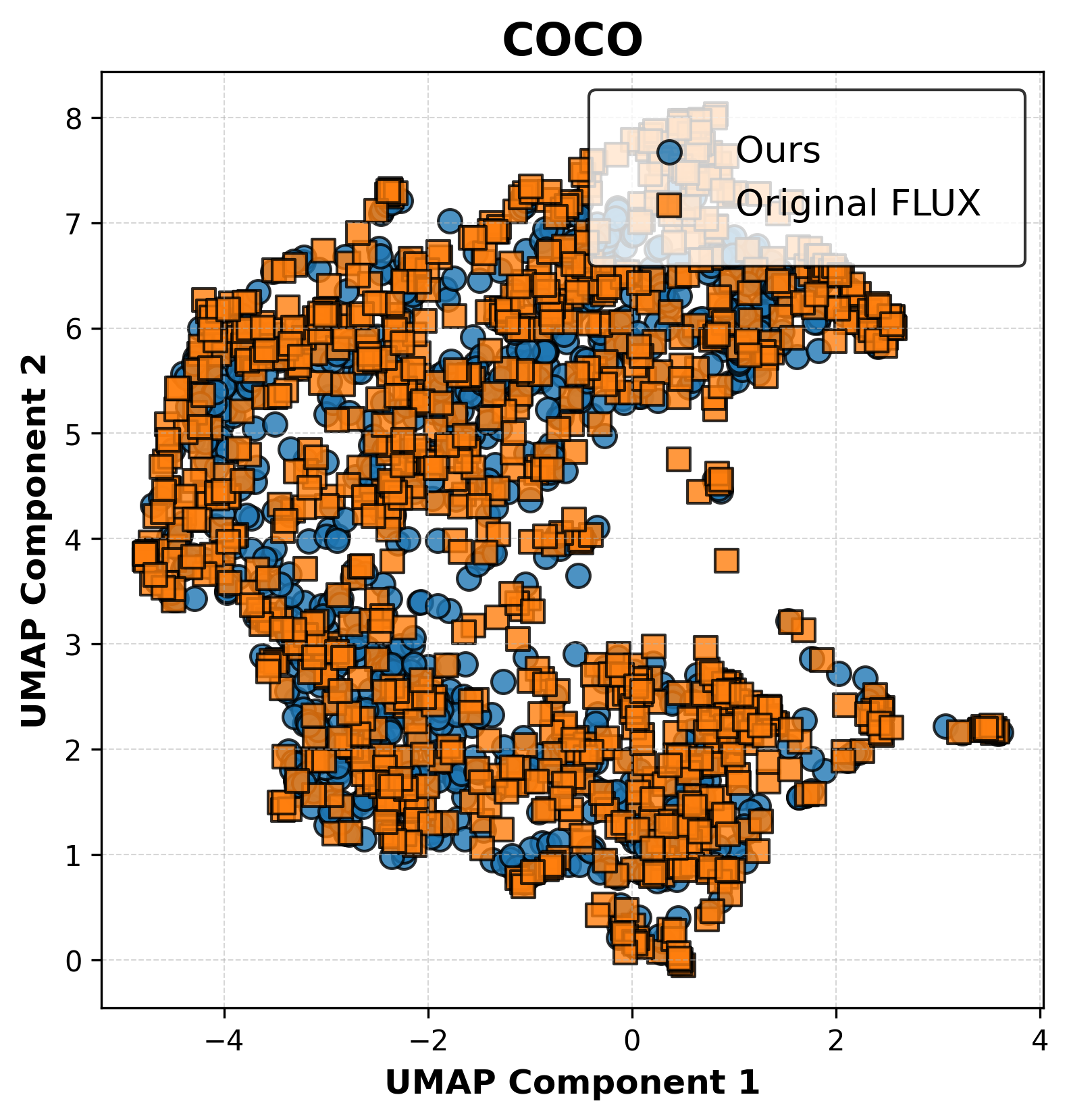}
        \caption{Neutral Prompts: COCO}
        \label{fig:coco}
    \end{subfigure}
    \caption{Visualizing the embedding space with UMAP. For this purpose, we select a subset of samples and project the reshaped denoised latent $\mathbf{z}_N$ into a 2D embedding space using UMAP for dimensionality reduction. All the latent embeddings are generated via our \textbf{medium} \textbf{UniForget} model. All the blue circles represent the image latents from \textbf{UniForget}, while the orange squares represent the image latents from the FLUX model. Figure~\ref{fig:cele} and Figure~\ref{fig:comic} demonstrate that for copyrighted prompts, the image embedding space of our model can be effectively decoupled from the original embedding space. In contrast, Figure~\ref{fig:coco} shows that for neutral prompts, the pruned model embedding space is still largely overlapped with an original model, which validates our first hypothesis.}
    \label{fig:embedding_space}
\end{figure*}

\section{Experiments}
\label{sec:exp}

\subsection{Experiment Setup.} For training, we randomly select 256 neutral text prompts $\mathcal{C}_n$ from LAION 2B~\cite{schuhmann2022laion} using GPT-4o~\cite{achiam2023gpt} to ensure that the prompts do not contain any copyright-sensitive memorization content. To evaluate our approach, we conduct all experiments using SOTA models for text-to-image generation, FLUX schnell and FLUX dev. Detailed implementation of our experiments can be found in Appendix D.

\subsection{Evaluation Datasets and Metrics.} To assess the effectiveness of our method, we conduct experiments using CLIP~\cite{radford2021learning} and SOTA VLMs, GPT-4o and LLaVA 1.6~\cite{achiam2023gpt, liu2024llava}. Our evaluations are conducted on the datasets containing sensitive memorization content to evaluate the effectiveness of copyrighted de-memorization. Specifically, we use the Comic dataset, which includes 25K characters with potential copyright risks~\cite{fivethirtyeight_comic_characters}, the IMDB Top 1000 Movies dataset~\cite{inductiveanks2023imdb}, and the Anime and Celebrity datasets from the CPDM dataset~\cite{ma2024dataset}. To provide a comprehensive evaluation, we test our method on more than 3000 instances of copyrighted memorization content across the following categories: 100 prompts for Anime, 2000 for Comic, 1000 for Movie, and 700 for Celebrity. We use a simple VQA approach with VLMs to check whether memorized content remains in the generated images. We ask questions like \texttt{"Does the image contain any copyrighted content?"} and \texttt{"Does the image have any <CONCEPT> related features?"} to get responses that act as evaluation metrics, called $\mathrm{Acc_{\text{g}}}$ and $\mathrm{Acc_{l}}$. A high $\mathrm{Acc}$ means a higher detection of copyrighted content, while a low $\mathrm{Acc}$ suggests better removal of memorized content. Additionally, we use CLIP~\cite{radford2021learning} and the Fréchet Inception Distance (FID)~\cite{heusel2017gans} to assess the quality of the generated images. We conduct tests on subsets of the COCO, Flickr30k, and LAION datasets, each consisting of 5,000 randomly selected images~\cite{lin2014microsoft, plummer2015flickr30k, schuhmann2022laion}. The detailed evaluation procedures are described in Appendix E.

\subsection{Main Results} \textbf{Hypothesis One: Decoupled embedding space between neutral and copyrighted features}. We validate this hypothesis by evaluating the denoised latent $\mathbf{z}_N$ corresponding to each generated image embedding using the prompts from the evaluation datasets. For demonstration purposes, we utilize UMAP~\cite{mcinnes2018umap} to reduce the dimensionality of the denoised latent embeddings to two~\cite{mcinnes2018umap}. Figure~\ref{fig:embedding_space} presents the visualization of the embedding space for two copyright-violating categories, Celebrity and Comic, along with the COCO dataset, which contains only neutral text prompts. Figure~\ref{fig:cele} shows that, using the prompts from the Celebrity dataset, the embedding space of our generated images exhibits a significantly large separation from the embeddings of the original model. Although some overlap between the two embedding groups still exists, it is not substantial. Similar embedding space decoupling can also be observed in Figure~\ref{fig:comic} for the Comic dataset, with even less overlap. Figure~\ref{fig:coco} demonstrates that, for neutral prompts from COCO, the latent space generated by our model shares a similar distribution and largely overlaps with that of the original model. These results in Figure~\ref{fig:embedding_space} visually confirm our first hypothesis: the embedding spaces for memorized and neutral content are largely separable. Our method effectively pushes generations away from the "copyrighted" clusters while preserving the distribution for neutral concepts.

\subsubsection{Validation using Latent Magnitude.}
According to \citet{wen2024detecting}, the high magnitude of text-conditional noise indicates memorization. We also conduct a similar experiment to evaluate the norm of the reduced embedding space from COCO and the Celebrity dataset generated by the original and our pruned model (medium). As depicted in Figure~\ref{fig:kde}, the KDE distribution of embedding magnitudes from the copyrighted prompts with our pruned model is significantly closer to the neutral prompts' distribution. This indicates that the magnitude of the generated image embeddings is noticeably smaller than that of the original model. This observation further verifies the effectiveness of our method.
\begin{figure}
    \centering
    \includegraphics[width=\linewidth]{images/fig-5-magnitude.pdf}
    \caption{KDE distributions of embedding magnitude for the original model and UniForget. As for the neutral prompt, COCO, the distribution remains largely unchanged. In contrast, for copyrighted prompts from Celebrity, the original model produces significantly different magnitude distributions compared to neutral prompts. After applying UniForget, the distributions of copyrighted prompts shift substantially closer to the neutral distributions.}
    \label{fig:kde}
\end{figure}

\begin{table*}[t!]
\caption{Quantitative evaluation of removing memorized copyrighted content. We assess models at three de-memorization levels: weak, medium, and strong, and on four datasets: Anime, Comic, Movie, and Celebrity. The effectiveness of de-memorization is measured using $\mathrm{Acc_{g}}$ and $\mathrm{Acc_{l}}$, leveraging GPT-4o, and LLaVA 1.6 to detect copyrighted content in generated images. General generative performance is evaluated using FID and CLIP metrics on the subset of the COCO dataset (5K). The `w retrain` variants show that a fine-tuning stage can restore generative quality to the original level. This involves a slight trade-off, yet performance in copyright de-memorization remains strong. \textbf{Bold} indicates the best score.}
\resizebox{\textwidth}{!}{
\small
\begin{tabular}{lcccccccccc}
\toprule
\textbf{Dataset} & \multicolumn{2}{c}{\textbf{Anime}} & \multicolumn{2}{c}{\textbf{Comic}} & \multicolumn{2}{c}{\textbf{Movie}} & \multicolumn{2}{c}{\textbf{Celebrity}} & \multicolumn{2}{c}{\cellcolor{gray!20}\textbf{COCO}}\\
\textbf{Model} & $\mathrm{Acc_{g}}\downarrow$ & $\mathrm{Acc_{l}}\downarrow$ & $\mathrm{Acc_{g}}\downarrow$ & $\mathrm{Acc_{l}}\downarrow$ & $\mathrm{Acc_{g}}\downarrow$ & $\mathrm{Acc_{l}}\downarrow$ & $\mathrm{Acc_{g}}\downarrow$ & $\mathrm{Acc_{l}}\downarrow$ & \cellcolor{gray!20}FID $\downarrow$ & \cellcolor{gray!20}CLIP $\uparrow$ \\
\cmidrule(lr){1-1} \cmidrule(lr){2-3} \cmidrule(lr){4-5} \cmidrule(lr){6-7} \cmidrule(lr){8-9} \cmidrule(lr){10-11}
FLUX-dev & 78.5\% & 81.3\% & 48.3\% & 87.1\% & 43.2\% & 36.5\% & 51.1\% & 55.4\% & \cellcolor{gray!20}\textbf{31.5} & \cellcolor{gray!20}\textbf{0.30} \\
FLUX-schnell & 68.7\% & 77.6\% & 42.6\% & 90.5\% & 37.1\% & 35.1\% & 43.2\% & 52.6\% & \cellcolor{gray!20}32.0 & \cellcolor{gray!20}0.29 \\
Our (weak) & 36.5\% & 52.2\% & 28.9\% & 70.2\% & 31.7\% & 26.3\% & 32.1\% & 36.7\% & \cellcolor{gray!20}32.1 & \cellcolor{gray!20}0.29 \\
Our (medium) & 30.3\% & 42.8\% & 10.2\% & 41.5\% & 23.3\% & 15.1\% & 17.4\% & 29.3\% & \cellcolor{gray!20}32.4 & \cellcolor{gray!20}0.28 \\
Our (medium) w retrain & 32.1\% & 44.5\% & 12.3\% & 43.9\% & 24.8\% & 16.9\% & 19.1\% & 31.2\% & \cellcolor{gray!20}31.8 & \cellcolor{gray!20}0.29 \\
Our (strong) & \textbf{16.3}\% & \textbf{27.4}\% & \textbf{9.1\%} & \textbf{33.3}\% & \textbf{16.6}\% & \textbf{13.4\%} & \textbf{5.3\%} & \textbf{23.3}\% & \cellcolor{gray!20}38.9 & \cellcolor{gray!20}0.25 \\
Our (strong) w retrain & 18.2\% & 29.8\% & 11.2\% & 35.1\% & 18.5\% & 15.2\% & 7.4\% & 25.6\% & \cellcolor{gray!20}\textbf{31.5} & \cellcolor{gray!20}0.29 \\
\bottomrule
\end{tabular}
}
\label{tab:concept_removal_eval}
\end{table*}

\noindent\textbf{Hypothesis Two: Selective parameters responsible for memorization}. We can significantly mitigate copyrighted memorization problems by removing small parts of the model. To validate this hypothesis, we conduct experiments as shown in Table~\ref{tab:concept_removal_eval} and Figure~\ref{fig:demo_results}. Each evaluation dataset consists solely of copyrighted prompts, where a lower detection accuracy indicates better de-memorization of copyrighted content. The results show that the baseline model, FLUX dev, and FLUX schnell exhibit significant copyrighted memorization issues, displaying high detection ratios across all categories. In contrast, our method substantially reduces the copyright detection rates in all categories, even when using the weak de-memorization model. Additionally, we assess our approach with LLaVA 1.6 and observe a similarly significant decline in detection rates. To evaluate the general generative performance, we use the COCO dataset, where FID and CLIP scores show a slight decrease compared to the original models. However, this reduction in performance is minimal relative to the substantial improvements in de-memorization. The overall results in Table~\ref{tab:concept_removal_eval} and Table~\ref{tab:masking_level} validate the hypothesis that removing a small specific part of model components is sufficient to mitigate memorization problems, including copyrighted content. Interestingly, with weak de-memorization, the quality of the generated images is noticeably better than the original images for prompts such as \texttt{"A photo of Lionel Messi"}. We also observe that for copyrighted prompts, the generated images by the original model tend to be noticeably blurrier. This further explains why the weak de-memorization model achieves an even better FID score than the original model, as shown in \mbox{Table~\ref{tab:general_quality_eval_updated}}.

\begin{table}[t]
\caption{Compatibility evaluation with CA~\cite{gandikota2023erasing} and UniForget. We use the unlearning method, CA, to remove the \texttt{"Nudity"} concept, and a lower score indicates better results.}
\centering
\resizebox{\columnwidth}{!}{
\small
\begin{tabular}{@{}lccccc@{}}
\toprule
Dataset          & \multicolumn{1}{c}{\begin{tabular}[c]{@{}c@{}}FLUX\\schnell \end{tabular}}& CA & \multicolumn{1}{c}{\begin{tabular}[c]{@{}c@{}}CA + UniForget\\(medium) \end{tabular}} & \multicolumn{1}{c}{\begin{tabular}[c]{@{}c@{}}CA + UniForget\\ (strong)\end{tabular}} \\ \midrule
I2P          &  30.3\%   &  \textbf{17.4}\%  &         18.1\%                &      19.3\%                   \\
P4D          &  51.6\%    &  22.3\%  &         \textbf{21.2}\%                &    25.3\%                     \\ \bottomrule
\end{tabular}}
\label{tab:quan_quality_eval}
\end{table}

\begin{table}[t]
\caption{Additional generative studies on neural prompts.}
\centering
\resizebox{\columnwidth}{!}{
\tiny
\begin{tabular}{@{}lcccc@{}}
\toprule
Dataset        & \multicolumn{2}{c}{LAION} & \multicolumn{2}{c}{Flickr} \\ \midrule
Metric        & FID  $\downarrow$        & CLIP  $\uparrow$       & FID  $\downarrow$        & CLIP $\uparrow$          \\
FLUX schnell &   \textbf{42.1}   &  \textbf{0.27}  &  40.1   &   0.32           \\
Our (weak)    &   47.4   &  0.25  &  \textbf{39.4}   &   \textbf{0.33}           \\
Our (medium)  &   50.9   &  0.23  &  40.2   &   0.32           \\
Our (strong)  &   62.1   &  0.21  &  48.7   &   0.27           \\ \bottomrule
\end{tabular}
}
\label{tab:general_quality_eval_updated}
\end{table}

\begin{figure}[t]
    \centering
    \includegraphics[width=\linewidth]{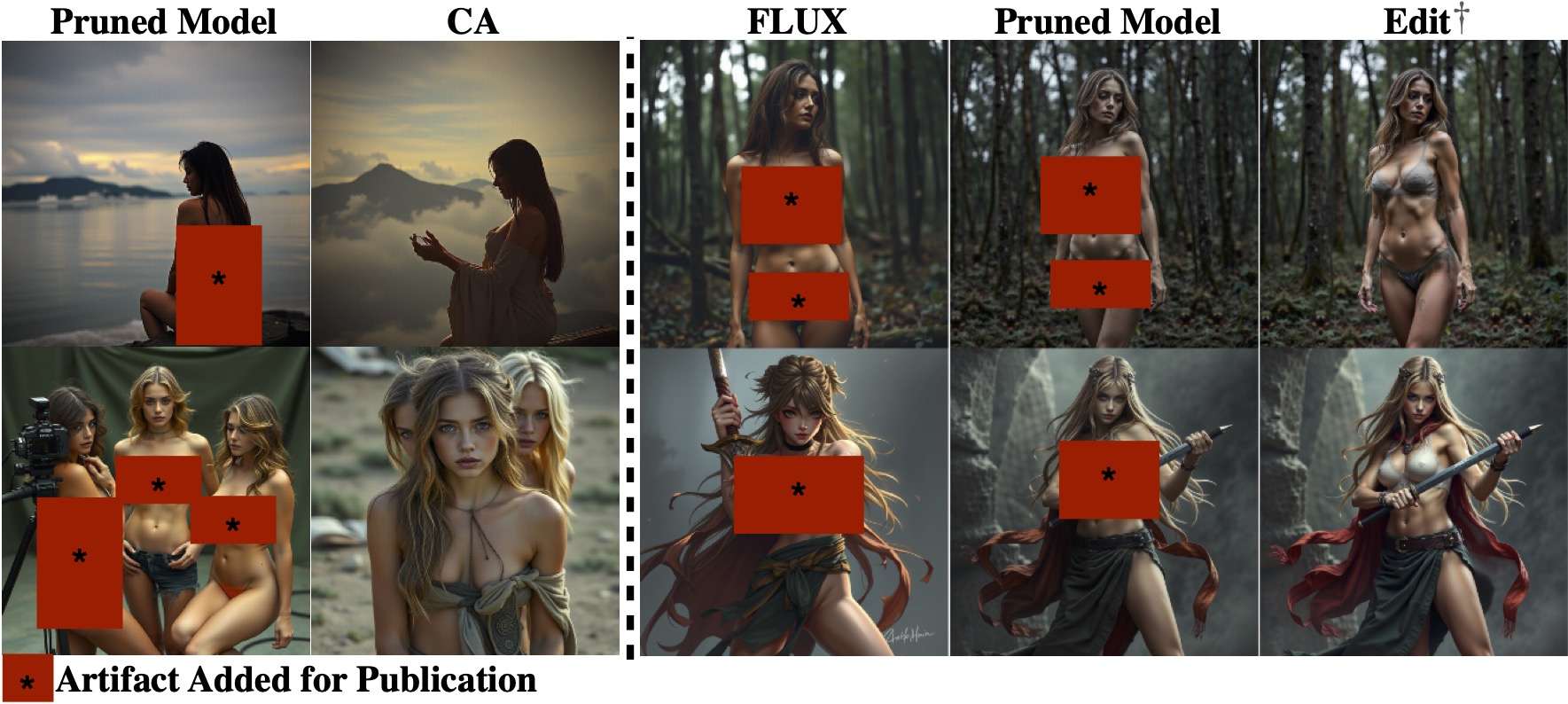}
    \caption{Compatibility experiment with the unlearning method CA~\cite{kumari2023ablating} and editing method~\cite{kulikov2024flowedit}. $\dagger$ indicates that we combined our pruned model with the editing method.}
    \label{fig:complementary}
\end{figure}
\subsection{Compatibility Studies}

\begin{figure}[t]
    \centering
    \includegraphics[width=\linewidth]{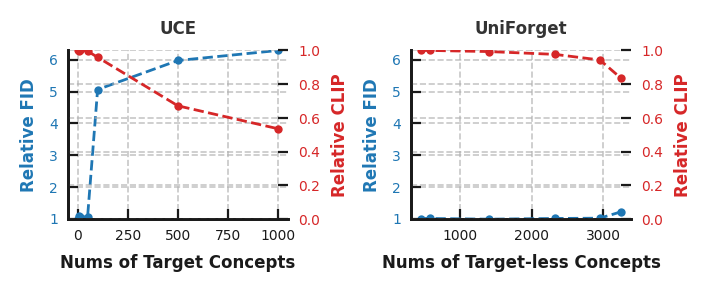}
    \caption{Unconscious concepts removing (Uniforget) via de-memorization vs. target concepts removing (UCE)~\cite{gandikota2024unified}. This Figure shows the relative FID and CLIP score comparison with the original model.}
    \label{fig:baseline_comparision}
\end{figure}

One of the main advantages of our method is that it is orthogonal to existing approaches in both the de-memorization and unlearning domains. As shown in \mbox{Figure~\ref{fig:baseline_comparision}}, the limitations of target concept unlearning only become significant when the number of target concepts is very high. In contrast, our method, UniForget, exhibits less performance degradation even in such cases. However, it cannot guarantee the complete removal of a specific concept. By combining our method with current unlearning techniques, we can effectively complement the strengths of each approach. Figure~\ref{fig:complementary} shows that Uniforget treats \texttt{"Nudity"} as a neutral concept and is unable to remove it. However, nudity content can be significantly reduced by applying target concept removal methods such as CA~\cite{kumari2023ablating}. Combining our model with the sampling process manipulation method ~\cite{kulikov2024flowedit} achieves remarkable performance in removing nudity concepts while mitigating general memorization problems. Additionally, we conduct a quantitative analysis to assess the impact of pruned models on the concept removal method CA~\cite{gandikota2023erasing}. Table~\ref{tab:quan_quality_eval} shows that the results of concept removal using pruned models are comparable to those obtained with original model. We also investigate the restoration of our pruned model's generative capabilities through full-parameter fine-tuning, utilizing the same training configuration as the de-memorization process. As detailed in Table~\ref{tab:concept_removal_eval}, this retraining process significantly improves both FID and CLIP scores over the non-retrained model. Crucially, the fine-tuned model preserves its strong de-memorization performance, underscoring our method's efficacy. This outcome indicates that sensitive content generation is localized to a small set of neurons, whose removal does not permanently compromise the model's broader generative potential.

\subsection{Ablation Studies} Table~\ref{tab:module_quality_eval} demonstrates an ablation study evaluating which part of the modules is essential for memorization. We observe that it is crucial to apply $\mathcal{M}_{\mathrm{ffn}}$ and $\mathcal{M}_{\mathrm{norm}}$ to ensure the removal of copyright-sensitive memory. We also assess the pruning ratio under different regularization parameters $\beta$, as presented in Table~\ref{tab:masking_level}, which supports the hypothesis that effective copyrighted de-memorization requires the removal of only a small subset of the model.

\begin{table}[t!]
\caption{Ablation study on \textbf{medium} UniForget across different modules of the FLUX model, evaluating general generative performance on COCO 5K and de-memorization effectiveness on the \textit{Anime} dataset. \textbf{Bold} indicates the best score.}
\resizebox{\columnwidth}{!}{
\small
\begin{tabular}{@{}ccccccc@{}}
\toprule
$\mathcal{M}_{\mathrm{attn}}$ & $\mathcal{M}_{\mathrm{ffn}}$ & $\mathcal{M}_{\mathrm{norm}}$  & $\mathrm{Acc_{g}} \downarrow$ & CLIP $\uparrow$ & FID $\downarrow$     \\ \midrule
\xmark               &  \xmark             & \xmark                & 77.6\%         & \textbf{0.29}   & \textbf{32.0} \\
\xmark               &  \checkmark         & \xmark                & 21.4\%         & 0.28   & 35.7  \\ 
\checkmark           &  \xmark             & \checkmark            & 63.5\%         & 0.21   & 45.4 \\ 
\xmark               &  \checkmark         & \checkmark            & \textbf{16.3\%}& 0.28   & 32.4 \\ \bottomrule
\end{tabular}
}
\label{tab:module_quality_eval}
\end{table}

\begin{table}[t]
\centering
\caption{Deactivation ratios of $\mathcal{M}_{\mathrm{ffn}}$ and $\mathcal{M}_{\mathrm{norm}}$ and the whole model's mask $\mathcal{M}$. By deactivating 16.3\% of the model, our work can largely mitigate the copyrighted model memorization.}
\small
\begin{tabular}{@{}lccc@{}}
\toprule
$\beta$ & $\mathcal{M}_{\mathrm{ffn}}$ & $\mathcal{M}_{\mathrm{norm}}$ & $\mathcal{M}$ \\ \midrule
$\beta_{\mathrm{weak}}$             &  24.2\%   &   5.6\%   &    6.7\%     \\
$\beta_{\mathrm{medium}}$           &  35.0\%   &    12.3\%  &   9.5\% \\
$\beta_{\mathrm{strong}}$            & 45.3\%  &  19.1\%   &   16.3\%      \\ \bottomrule
\end{tabular}
\label{tab:masking_level}
\end{table}

\section{Limitations and Future Work}
\label{sec:limi}

Although our method introduces a new perspective and achieves remarkable results, it involves certain trade-offs. \mbox{Figure 4} in Appendix H illustrates that our strong de-memorization model can also forget some neutral concepts. However, as shown in Table~\ref{tab:quan_quality_eval}, this performance drop is largely recovered after a brief retraining phase. Additionally, our evaluation relies on the performance of VLMs such as GPT-4o and LLaVA~\cite{achiam2023gpt, liu2023flow}. We observed that these models can exhibit variability in detecting copyrighted content, sometimes misclassify copyright-free content, or produce inconsistent judgments across different VLMs, as shown in Figure 4 in Appendix H. However, identifying copyright violations is challenging for both VLMs and humans, underscoring the complexity of this issue. Developing reliable methods for detecting copyright-sensitive content is an essential direction for the field. While this challenge is beyond the scope of our current research, we plan to explore it further in future work. We also aim to extend this work to other generative models, such as SD3 and the Visual Autoregressive Models~\cite{tian2024visual, rombach2022high}.

\section{Conclusion}
\label{sec:con}

This work studies and discusses the root cause of the memorization problem in generative models and formulates two essential hypotheses. Our proposed model validates these hypotheses and demonstrates that pruning a small subset of the model can largely mitigate the risk of generating copyrighted content. We further interpret the problem of model memorization through the embedding space, showing that our method effectively decouples the embedding subspace of copyrighted content from the embedding space of neutral prompts. Our work provides a new perspective on model copyrighted memorization and a simple yet effective solution. Additionally, our method is highly compatible with existing approaches, and we believe that it can significantly advance de-memorized and copyright-compliant generative models and contribute to the development of responsible generative \mbox{models}. 

\bibliography{aaai2026}

\fi
\setlength{\leftmargini}{20pt}
\makeatletter\def\@listi{\leftmargin\leftmargini \topsep .5em \parsep .5em \itemsep .5em}
\def\@listii{\leftmargin\leftmarginii \labelwidth\leftmarginii \advance\labelwidth-\labelsep \topsep .4em \parsep .4em \itemsep .4em}
\def\@listiii{\leftmargin\leftmarginiii \labelwidth\leftmarginiii \advance\labelwidth-\labelsep \topsep .4em \parsep .4em \itemsep .4em}\makeatother

\setcounter{secnumdepth}{0}
\renewcommand\thesubsection{\arabic{subsection}}
\renewcommand\labelenumi{\thesubsection.\arabic{enumi}}

\newcounter{checksubsection}
\newcounter{checkitem}[checksubsection]

\newcommand{\checksubsection}[1]{%
  \refstepcounter{checksubsection}%
  \paragraph{\arabic{checksubsection}. #1}%
  \setcounter{checkitem}{0}%
}

\newcommand{\checkitem}{%
  \refstepcounter{checkitem}%
  \item[\arabic{checksubsection}.\arabic{checkitem}.]%
}
\newcommand{\question}[2]{\normalcolor\checkitem #1 #2 \color{blue}}
\newcommand{\ifyespoints}[1]{\makebox[0pt][l]{\hspace{-15pt}\normalcolor #1}}

\section*{Reproducibility Checklist}

% The questions start here

\checksubsection{General Paper Structure}
\begin{itemize}

\question{Includes a conceptual outline and/or pseudocode description of AI methods introduced}{(yes/partial/no/NA)}
partial

\question{Clearly delineates statements that are opinions, hypothesis, and speculation from objective facts and results}{(yes/no)}
yes

\question{Provides well-marked pedagogical references for less-familiar readers to gain background necessary to replicate the paper}{(yes/no)}
yes

\end{itemize}
\checksubsection{Theoretical Contributions}
\begin{itemize}

\question{Does this paper make theoretical contributions?}{(yes/no)}
no

	\ifyespoints{\vspace{1.2em}If yes, please address the following points:}
        \begin{itemize}
	
	\question{All assumptions and restrictions are stated clearly and formally}{(yes/partial/no)}
	NA

	\question{All novel claims are stated formally (e.g., in theorem statements)}{(yes/partial/no)}
	NA

	\question{Proofs of all novel claims are included}{(yes/partial/no)}
	NA

	\question{Proof sketches or intuitions are given for complex and/or novel results}{(yes/partial/no)}
	NA

	\question{Appropriate citations to theoretical tools used are given}{(yes/partial/no)}
	NA

	\question{All theoretical claims are demonstrated empirically to hold}{(yes/partial/no/NA)}
	NA

	\question{All experimental code used to eliminate or disprove claims is included}{(yes/no/NA)}
	NA
	
	\end{itemize}
\end{itemize}

\checksubsection{Dataset Usage}
\begin{itemize}

\question{Does this paper rely on one or more datasets?}{(yes/no)}
yes

\ifyespoints{If yes, please address the following points:}
\begin{itemize}

	\question{A motivation is given for why the experiments are conducted on the selected datasets}{(yes/partial/no/NA)}
	yes

	\question{All novel datasets introduced in this paper are included in a data appendix}{(yes/partial/no/NA)}
	yes

	\question{All novel datasets introduced in this paper will be made publicly available upon publication of the paper with a license that allows free usage for research purposes}{(yes/partial/no/NA)}
	yes

	\question{All datasets drawn from the existing literature (potentially including authors' own previously published work) are accompanied by appropriate citations}{(yes/no/NA)}
	yes

	\question{All datasets drawn from the existing literature (potentially including authors' own previously published work) are publicly available}{(yes/partial/no/NA)}
	yes

	\question{All datasets that are not publicly available are described in detail, with explanation why publicly available alternatives are not scientifically satisficing}{(yes/partial/no/NA)}
	NA

\end{itemize}
\end{itemize}

\checksubsection{Computational Experiments}
\begin{itemize}

\question{Does this paper include computational experiments?}{(yes/no)}
yes

\ifyespoints{If yes, please address the following points:}
\begin{itemize}

	\question{This paper states the number and range of values tried per (hyper-) parameter during development of the paper, along with the criterion used for selecting the final parameter setting}{(yes/partial/no/NA)}
	yes

	\question{Any code required for pre-processing data is included in the appendix}{(yes/partial/no)}
	yes

	\question{All source code required for conducting and analyzing the experiments is included in a code appendix}{(yes/partial/no)}
	yes

	\question{All source code required for conducting and analyzing the experiments will be made publicly available upon publication of the paper with a license that allows free usage for research purposes}{(yes/partial/no)}
	yes
        
	\question{All source code implementing new methods have comments detailing the implementation, with references to the paper where each step comes from}{(yes/partial/no)}
	yes

	\question{If an algorithm depends on randomness, then the method used for setting seeds is described in a way sufficient to allow replication of results}{(yes/partial/no/NA)}
	partial

	\question{This paper specifies the computing infrastructure used for running experiments (hardware and software), including GPU/CPU models; amount of memory; operating system; names and versions of relevant software libraries and frameworks}{(yes/partial/no)}
	yes

	\question{This paper formally describes evaluation metrics used and explains the motivation for choosing these metrics}{(yes/partial/no)}
	yes

	\question{This paper states the number of algorithm runs used to compute each reported result}{(yes/no)}
    yes

	\question{Analysis of experiments goes beyond single-dimensional summaries of performance (e.g., average; median) to include measures of variation, confidence, or other distributional information}{(yes/no)}
	yes

	\question{The significance of any improvement or decrease in performance is judged using appropriate statistical tests (e.g., Wilcoxon signed-rank)}{(yes/partial/no)}
	partial

	\question{This paper lists all final (hyper-)parameters used for each model/algorithm in the paper’s experiments}{(yes/partial/no/NA)}
	yes

\end{itemize}
\end{itemize}
\ifreproStandalone
\end{document}